\documentclass{article}

\usepackage[final]{mlsafety_neurips_2022}

\usepackage[utf8]{inputenc} %
\usepackage[T1]{fontenc}    %
\usepackage{hyperref}       %
\usepackage{url}            %
\usepackage{booktabs}       %
\usepackage{amsfonts}       %
\usepackage{nicefrac}       %
\usepackage{microtype}      %
\usepackage{xcolor}         %
\usepackage{amsmath,amsfonts,bm,amsthm}

\theoremstyle{theorem}

\newtheorem{claim}[]{Claim}

\def\eqref#1{equation~\ref{#1}}

\def\1{\bm{1}}

\DeclareMathAlphabet{\mathsfit}{\encodingdefault}{\sfdefault}{m}{sl}
\SetMathAlphabet{\mathsfit}{bold}{\encodingdefault}{\sfdefault}{bx}{n}

\usepackage{mathtools}
\usepackage{caption}
\usepackage{subcaption}

\usepackage{todonotes}

\title{Unifying Grokking and Double Descent}

\author{Xander Davies\thanks{Equal contribution. Correspondence to \texttt{alexander\_davies@college.harvard.edu}.}  \\
Harvard University
\And
Lauro Langosco\footnotemark[1] \\
University of Cambridge
\And 
David Krueger \\
University of Cambridge
}

\theoremstyle{definition}

\begin{document}

\maketitle

\begin{abstract}

A principled understanding of generalization in deep learning may require unifying disparate observations under a single conceptual framework.
Previous work has studied \emph{grokking}, a training dynamic in which a sustained period of near-perfect training performance and near-chance test performance is eventually followed by generalization, as well as the superficially similar \emph{double descent}. These topics have so far been studied in isolation. We hypothesize that grokking and double descent can be understood as instances of the same learning dynamics within a framework of pattern learning speeds. We propose that this framework also applies when varying model capacity instead of optimization steps, and provide the first demonstration of model-wise grokking.

\end{abstract}

\section{Introduction}

\begin{figure}[h]
\centering
\begin{subfigure}[t]{0.32\textwidth}
 \centering
 \includegraphics[width=\textwidth]{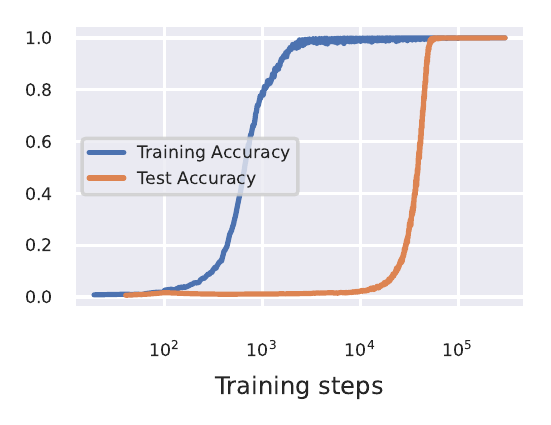}
 \caption{Typical grokking on a modular division task.}
 \label{fig:grokking-acc}
\end{subfigure}
\hfill
\begin{subfigure}[t]{0.32\textwidth}
 \centering
 \includegraphics[width=\textwidth]{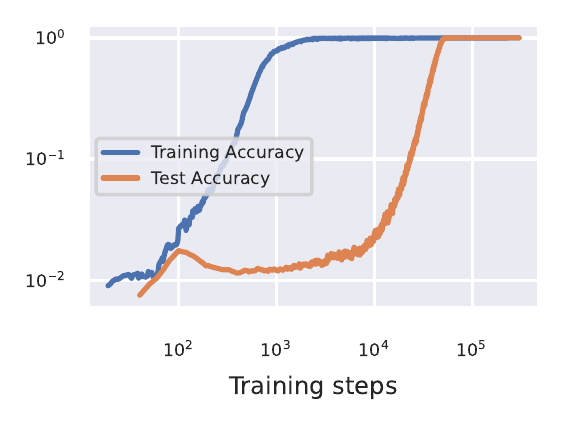}
 \caption{Plotting log accuracy reveals a small double descent.}
 \label{fig:grokking-log-acc}
\end{subfigure}
\hfill
\begin{subfigure}[t]{0.32\textwidth}
 \centering
 \includegraphics[width=\textwidth]{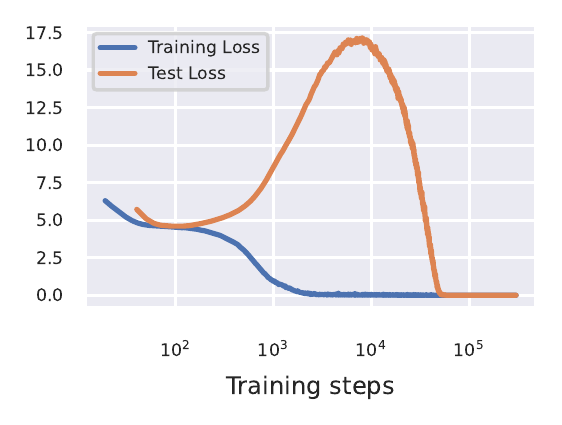}
 \caption{The double descent is more pronounced in loss.}
 \label{fig:grokking-loss}
\end{subfigure}
\caption{
    Three views of the same training run, replicated from \citet{power2022grokking}.
}
\label{fig:epoch-wise-grokking}
\end{figure}

On some datasets, neural networks exhibit surprising training dynamics termed \emph{grokking} by \citet{power2022grokking}. In grokking, the model initially overfits, achieving perfect performance on the training set while remaining at near-chance performance on the test set. Later in training, test performance improves and the model eventually achieves perfect test accuracy. 
Grokking is reminiscent of the \textit{double descent} phenomenon \citep{belkin2018reconciling, nakkiran2021deep}, in which test performance initially improves, then worsens as the model overfits, and then eventually improves again as model capacity increases.

We argue that double descent and grokking are best viewed as two instances of the same phenomenon,
in which inductive biases prefer better-generalizing but slower to learn patterns, leading to a transition from poorly-generalizing to well-generalizing patterns. This transition happens both \emph{epoch-wise} as a function of training time (Figure~\ref{fig:epoch-wise-grokking}) and \emph{model-wise} as a function of model size (Figure~\ref{fig:heatmap-loss}).

\paragraph{Contributions.}

We provide a conceptual framework (Claim~\ref{hyp:pattern}, Claim~\ref{hyp:model-wise}) which unifies the learning dynamics of grokking and double descent. We propose a simple model (Section~\ref{sec:equations}) to support Claim~\ref{hyp:pattern}, and provide evidence for Claim~\ref{hyp:model-wise} by showing that grokking can occur as a function of model size (Section~\ref{sec:experiments}).

\begin{figure}[t]
\centering
\includegraphics[width=\textwidth]{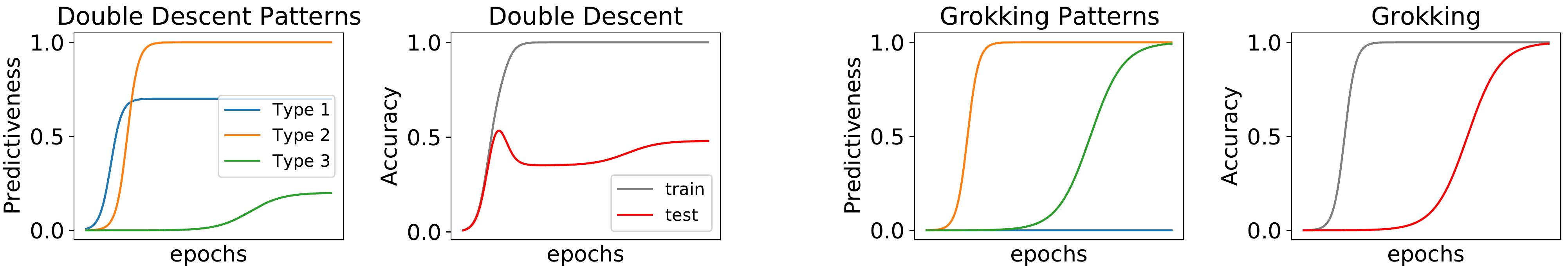}
\caption{
    \textbf{A simple model of pattern learning.} Left to right, (1) development of pattern predictiveness during double descent, with well-performing Type 1 (heuristics) patterns and somewhat-predictive Type 3 (slow \& well generalizing) patterns ; (2) resulting train and test accuracy curves for double descent; (3) pattern predictiveness during grokking, with poorly-performing Type 1 patterns and perfectly-predictive Type 3 patterns; (4) resulting train and test curves for grokking. In both cases, Type 2 (medium \& poorly-generalizing) patterns quickly achieve perfect predictiveness.
}
\label{fig:toy}
\end{figure}

\section{Pattern Learning} \label{sec:pattern-learning}
We propose an abstract model of neural network training as pattern learning. Our model aims to capture the intuitive notion that a network uses distinct kinds of mechanisms to classify different inputs; for example, it might memorize some examples while applying heuristics to classify others.
We build on this notion of pattern to formulate our Claims~\ref{hyp:pattern}~and~\ref{hyp:model-wise}, and describe our model in more detail in Section~\ref{sec:equations}.

\subsection{Core Claims} \label{subsec:claims}
We first claim that epoch-wise double descent and grokking share the same underlying dynamics, and then claim that these dynamics also transfer to the model-wise setting.

\begin{claim}[Pattern learning dynamics] \label{hyp:pattern} Grokking, like epoch-wise double descent, occurs when slow patterns generalize well and are ultimately favored by the training regime, but are preceded by faster patterns which generalize poorly.
\end{claim}

We formalize Claim~\ref{hyp:pattern} in Section~\ref{sec:equations}, where we propose a simple model of pattern learning to explain neural network training dynamics. As shown in Figure~\ref{fig:toy}, our model can produce the dynamics from grokking as well as double descent.

\begin{claim}[Duality between model size and scaling time] \label{hyp:model-wise}
Patterns can be viewed as a function of model size as well as a function of training time, and learning dynamics are similar between these two cases.\footnote{%
It is reasonable to speak of pattern learning as a function of effective capacity \citep{arpit2017closer} or effective model complexity \citep{nakkiran2021deep}, though we avoid doing so in this work so as to stay agnostic about the precise metric.} %
\end{claim}

If Claim~\ref{hyp:model-wise} is correct, then grokking, like double descent, should happen model-wise as well as epoch-wise. We confirm this hypothesis empirically in Section~\ref{subsec:model-wise}.

\subsection{A Model of Learning Dynamics} \label{sec:equations}

In our model, a neural network consists of a set of $n$ patterns that develop over time during training.
For any training point, the probability that pattern $i$ classifies it correctly at time-step $t$ is $p_i(t)$.%
We model $p_i$ as a sigmoid
\begin{equation*}
p_i(t) = \frac{\gamma_i}{1 + e^{-\alpha_i(t - b_i)}},
\end{equation*}
parameterized by a maximum predictiveness $0 \leq \gamma_i \leq 1$, inflection point $b_i \geq 0$, and learning speed $\alpha_i \geq 0$.
The probability that the overall model classifies the training point correctly is then given by the probability that any of the patterns does so. 
Since each data-point is classified independently, the probability of classifying a single training point correctly is equal to the overall training accuracy,
\begin{align} \label{eq:train-acc}
\text{acc}_{\text{train}} (t) = 1 - \prod_{i=1}^n \left(1 - p_i(t) \right).
\end{align}
To calculate test accuracy of our model we assign additional generalization parameters $g_1, \dots, g_n \in [0, 1]$ to represent the idea that some patterns generalize better than others.\footnote{
In this model whether the network classifies a given input correctly depends on whether it is in the training set or the test set, which does not make sense in the case of i.i.d. data: in the real world, whether a given example $(x,y)$ is classified correctly or not is purely a fact about the value of $(x,y)$ and the parameters of the model, and doesn't depend on whether its in the training or test set.
However, we include such a dependence to model neural networks that overfit the training set and thus are better able to classify training examples.
Intuitively, every pattern corresponds to a `generalization set' of inputs that are classified using that pattern even if they did not occur during training. The generalization parameter can then be viewed as the fraction of the generalization set that is classified correctly.
}
We assume that patterns that are more frequently successful during training are more used at test time. For any subset $A$ of $\{1, \dots, n\}$, we define the probability $P_A(t) = \prod_{i \in A}p_i(t) \prod_{j \notin A} (1 - p_j(t))$ that only the patterns in $A$ are successful, as well as the average generalization ability $G(A) = 1/{\vert A \vert} \sum_{i \in A} g_i$ of the patterns in $A$. Overall test accuracy at time $t$ is then given by the average generalization ability weighted the probability of success at train time,
\begin{equation} \label{eq:test-acc}
    \text{acc}_\text{test}(t) = \sum_{A \subseteq \{1, \dots, n\}} P_A(t) G(A).
\end{equation}
We additionally introduce the notion of a \textit{preferred pattern}, which we use to model the effect of inductive biases on pattern development.
When pattern $k$ is preferred, $g_k$ is weighted more highly in Equation~\ref{eq:test-acc} by setting $G(A) = g_k$ if $k \in A$ and otherwise $G(A) = 1/\vert A \vert \sum_{i \in A} g_i$ as usual.

\begin{figure}[t]
\centering
\includegraphics[width=\textwidth]{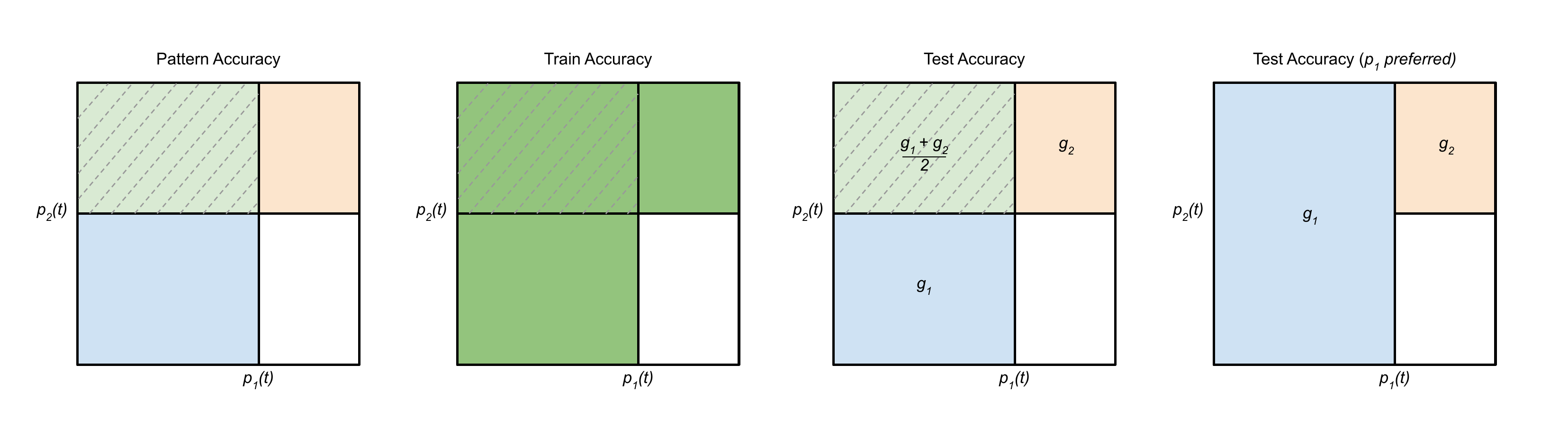}
\caption{
    \textbf{Two pattern model.}
   Full square represents the data set. Left to right, (1) blue shows training samples correctly classified by $p_1(t)$, red shows training samples correctly classified by $p_2(t)$, dashed green shows samples correctly classified by both; (2) accuracy is the union of patterns 1 and 2; (3) test accuracy is the sum of correct classifications weighted by generalization parameters, and by their average generalization in the case both correctly classifying; (4) test accuracy given pattern 1 is \textit{preferred}, causing all examples correctly classified by pattern 1 to generalize according to $g_1$.
}
\label{fig:patterns}
\end{figure}

\subsection{Double Descent \& Grokking with Pattern Types}

We model learning in double descent and grokking as interactions between three patterns learned at different speeds during training:
a \textbf{heuristic pattern} that is fast and generalizes well; an \textbf{overfitting pattern} that is fast, though slower than the heuristic, and generalizes poorly; and a \textbf{slow-generalizing pattern} that is slow and generalizes well. This last pattern is ultimately preferred by the inductive biases of the model.

As shown in Figure~\ref{fig:toy}, this simple model can produce grokking as well as double descent; the two can be interpolated between by solely modulating the maximum predictiveness ($\gamma$) of the heuristic and the slow-generalizing patterns.\footnote{Code replicating our results is available at \href{https://github.com/xanderdavies/unifying-grok-dd}{\texttt{github.com/xanderdavies/unifying-grok-dd}}, including an interactive notebook to explore the pattern learning model and interpolate between grokking and double descent.}

\section{Experiments} \label{sec:experiments}

\begin{figure}[t]
\centering
\includegraphics[width=\textwidth]{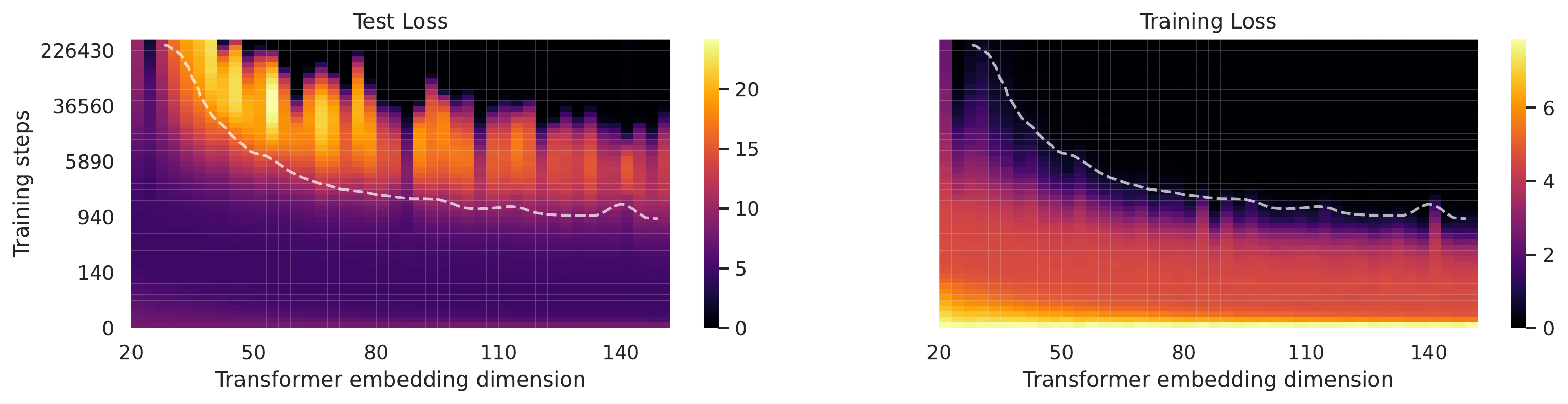}
\caption{
    \textbf{Model-Wise Grokking.} Darker color indicates lower loss, dotted line represents 90\% training accuracy.
   \textit{Left:} Test loss as function of training steps and model size. \textit{Right:} Training loss. Epoch-Wise Grokking: Moving vertically from zero training steps, test loss initially falls, then rises dramatically (beginning near perfect fitting of the training set), and then falls again. Model-Wise Grokking: Moving horizontally along training step counts which will eventually lead to low loss, we see the same pattern of initial test loss decrease, followed by rapid increase around training-set fitting, and then eventual decrease. See Figure \ref{fig:model-wise-grokking} for a horizontal cross-section.
}
\label{fig:heatmap-loss}
\end{figure}

\subsection{Model-Wise Grokking} \label{subsec:model-wise}

Following the original grokking setting of \cite{power2022grokking}, we train decoder-only transformers \citep{vaswani2017attention} with causal attention masking on the binary operation of division mod 97 (details in Appendix~\ref{sec:experimental-detail}). We explore the effects of regulating effective model complexity by varying parameter-count.

\begin{figure}[t]
\centering
\includegraphics[width=\textwidth]{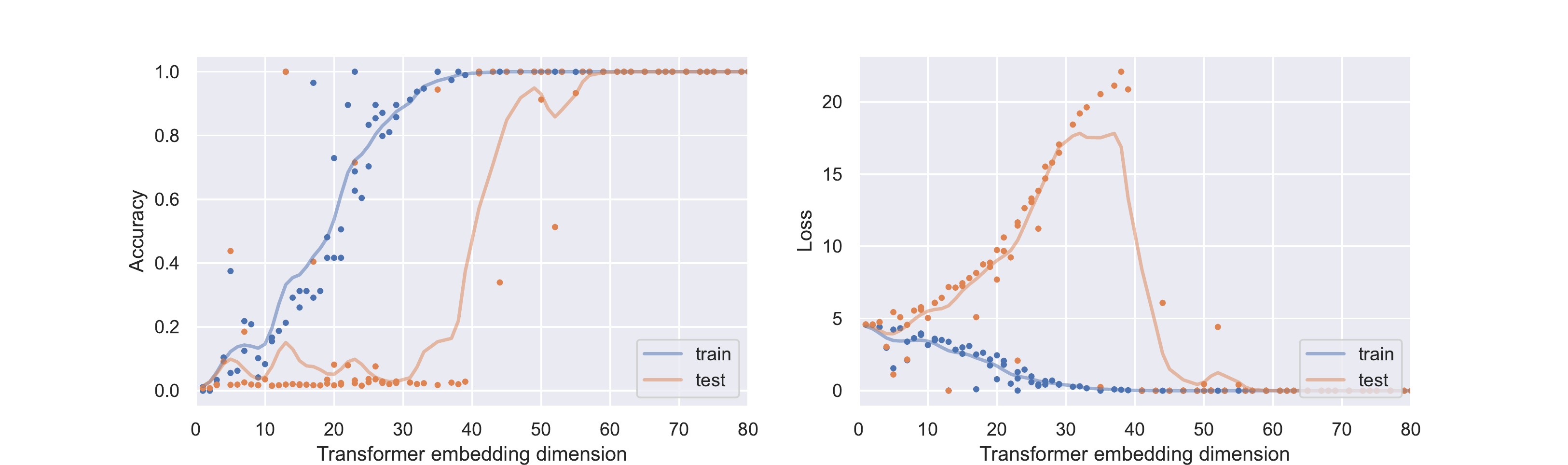}
\caption{
    Model-wise grokking, with each point a transformer of the corresponding embedding dimension trained for 400K optimization steps. Lines are smoothed data via a Gaussian filter. \textbf{Left:} Accuracy, with delayed generalization. \textbf{Right:} Loss, with a rise in test loss followed by a fast descent. 
}
\label{fig:model-wise-grokking}
\end{figure}

Figure~\ref{fig:heatmap-loss} shows train and test loss with respect to optimization steps and embedding dimension. For a range of values, delayed generalization occurs both when moving vertically (epoch-wise grokking), as well as when moving horizontally (model-wise grokking). Figure~\ref{fig:heatmap-acc} shows the corresponding accuracy heatmaps, with clear double descent behavior both epoch-wise and model-wise. Figure~\ref{fig:model-wise-grokking} shows model-wise grokking with models trained for 400k epochs.

\subsection{A Type 1 Pattern in the Grokking Setting} \label{subsec:initial-descent}
In the modular division setting of \citet{power2022grokking} the test accuracy is non-monotonic (Figure~\ref{fig:epoch-wise-grokking}). Early in training, the model learns that for all $b$, $0 / b = 0 \mod n$. This pattern generalizes well and is learned quickly, corresponding to a Type 1 pattern in our model. As predicted by our model, this leads to an initial spike above chance in test accuracy. Interestingly, the development of poorly-generalizing Type 2 features then leads to \textit{worse-than-chance} performance on the rest of the data, likely due to anti-correlation between train and test set values for every dividend and divisor in modular division, causing a descent in test performance. This small bump is noticeable in \citet{power2022grokking}, but not noted by the authors. We validate this hypothesis experimentally (Figure~\ref{fig:initial-ascent}).

\section{Related Work} 

\paragraph{Pattern Learning at Different Speeds.} 
\citet{heckel_early_2020} find different parts of networks learning at different speeds can cause two bias-variance trade-off curves with different minima, leading to double descent in test error; they show that adjusting step-sizes of different layers can align the learning curves and eliminate double descent behavior. They argue that epoch-wise double descent occurs due to this difference in learning speed, as opposed to due to regulating model complexity. \citet{pezeshki_multi-scale_2021} use a linear teacher-student model to demonstrate that epoch-wise double descent can be explained by different patterns being learned at different speeds. \citet{stephenson_when_2021} experimentally demonstrate double descent can be avoided by removing or accelerating slow-to-learn features. In this work, we extend this pattern learning framework to interpret grokking behavior as well as double descent behavior, and propose a mathematical model uniting both phenomena. Our framework uses the concept of distinct patterns with individual strengths \citep{heckel_early_2020, pezeshki_multi-scale_2021}, but is distinct in allowing for specific per-pattern generalization parameters and separating maximum predictiveness from learning speed, as well as introducing the notion of a ``preferred” pattern.

\paragraph{Double Descent.} The double descent phenomenon describes a phenomenon in which model test performance initially improves, then worsens as the model overfits, and then eventually improves again as we increase the \textit{capacity} of our training procedure \citep{belkin2018reconciling}. This capacity measure was originally demonstrated with model size (model-wise); \citet{nakkiran2021deep} generalize the capacity notion in \textit{effective model complexity}, and show double descent can occur when varying the number of optimization steps (epoch-wise) and other forms of capacity modulation. 

\paragraph{Grokking.} 
Grokking was first demonstrated by \citet{power2022grokking}, who state that grokking is distinct from double descent because 
generalization occurs far past the interpolation threshold.
\citet{liu2022towards} demonstrate that grokking generalization speed can be accelerated in a simple embedding-
decoder model by using a larger learning rate for the embedding than for the decoder.
This is in line with findings by \citet{heckel_early_2020} on epoch-wise double descent, where decreasing learning rates in later layers (which learn faster) aligns pattern learning speeds. We consider this further evidence that both grokking and epoch-wise double descent occur as a result of similar  learning dynamics resulting from different speeds of pattern development.
\citet{nanda_mechanistic_2022} investigate grokking through mechanistic interpretability, with findings in line with our results (specifically observing the development of a Type 3 pattern).

\paragraph{Weight Decay} \label{subsec:weight-decay}
Both \citet{nakkiran2021deep} and \citet{pezeshki_multi-scale_2021} find that weight decay acts as a capacity constraint, resulting in both weight decay-wise double descent and preventing the learning of slower features. In the grokking setting, however, weight decay plays a significant role in speeding up time to generalization \citep{power2022grokking}. We speculate that this occurs due to weight decay hindering the memorization-based solutions, thus accelerating the development of the better-generalizing solution.

\section{Conclusion}

In this work, we cover two phenomena in deep learning that were previously studied in isolation---grokking and double descent---and argue that they are best understood as instances of the same underlying learning dynamics.

The project of building a principled understanding of how and when networks generalize may be especially important from the point of view of \emph{AGI safety}, the problem of ensuring that advanced AI systems are safe despite pursuing misaligned goals \citep{carlsmith2022report,bostrom2014superintelligence} and incentives to seek power or deceive human operators for instrumental reasons \citep{turner2021power,omohundro2008}.
A central problem is that we may need to be certain of the safety of a model before we scale it to a capability level beyond which we cannot control it. 
Concerningly, it is well-known that the out-of-distribution (OOD) generalization behavior of deep learning systems can be hard to control or predict.
A robust theory of generalization and learning in neural networks may be necessary to solve this problem.

\section{Acknowledgements}
Thanks to Jesse Hoogland, Antonia Marcu, Dmitrii Krasheninnikov, Robert Kirk, Davis Brown, Peter Barnett, Rohin Shah, Nathan Bernard, Neel Nanda, Max Nadeau, Max Kaufmann, Alan Chan, Michaël Trazzi, and Oam Patel for valuable discussion, feedback on drafts of this paper and / or contributing suggestions for experiments.

\bibliography{refs}
\bibliographystyle{iclr2022_conference}

\newpage
\appendix

\section{An alternative intuition for preferred patterns}
\label{subsec:alt-formalism}

Rather than thinking about patterns as random functions from the entire input distribution $\mathcal{D}$ to the model's output $\mathcal{O}$, we can instead think of patterns as functions from a \textit{subset} of the input distribution $\mathcal{D}_i \subset \mathcal{D}$.
On this view, a pattern can successfully classify an input from the training set if and only if that input is in the patterns domain. On the test set, a data-point is allocated uniformly at random to patterns whose domain includes that data-point. If allocated to pattern $j$, the test data-point is classified successfully with probability $g_j$.

We can then say that a \textit{preferred pattern} enforces that its domain $\mathcal{D}_i$ is disjoint from all other pattern domains on the test set; that is $\mathcal{D}_i$ is preferred if $\forall j \in \{0, \cdots, n\}, i \neq j \implies \mathcal{D}_i \cap \mathcal{D}_j = \emptyset$.

\section{Experiment Details} \label{sec:experimental-detail}

In all of our experiments, we train decoder-only transformer with causal attention masking. 
Each residue is encoded as a symbol, and loss and accuracy are only evaluated on the answer part of the equation. Unless otherwise stated, we use a 2-layer network of width 128, with a single attention head. We typically train for 400 thousand optimization steps via AdamW ($\beta_1 = 0.9$, $\beta_2 = 0.98$), with learning rate of 1e-3 and weight decay of 1e-5.

\section{Discussion: how relevant is the science of deep learning for AI safety?}
A key factor that influences whether we will be able to build safe AI systems is the degree to which our systems are transparent and well-understood.
A system is transparent if we are able to ascertain what process lead it to produce a specific output or action; we might try to train a system to produce visible thoughts, truthfully report its knowledge, or probe it with interpretability tools \citep{arc2021elk,soares2021visible,lanham2022oversight,lin2021truthfulqa,olah2018interp}.
A system is well-understood to the extent that we have a broader theory about why it works and how it scales with respect to training time, model size, or dataset size. To name a few examples: we might try to quantitatively understand why models follow certain scaling laws, how they behave in the infinite-width limit, or what training curves look like on different input types
\citep{sharma2022manifold, bahri2021explaining, maloney2022solvable, yang2021tensor, siddiqui2022metadata}.

We use the term \emph{science of deep learning} to refer to the latter kind of inquiry,
which can be directly useful for safety: for example, it might help us predict when and how a neural network acquires a certain capability or behavior. This is helpful if we want to build a system that can perform a specific feat of engineering, while making sure that it does not have the capability to model humans; or if we tried to build a system that optimizes for the short-term approval of its operator, and would like to know in advance if it will generalize in a way that leads to it deceiving its operator in order to gain a higher measured approval score once it gains the ability to do so.
In addition, a better science of deep learning may help us build better interpretability tools. %

A science of deep learning may also be harmful if theoretical or empirical breakthroughs result in faster advancement of potentially dangerous capabilities. This is a major drawback of this line of research.
A tentative argument in favor of scientific / conceptual work is that it seems possible to build human-level AI systems without a radically improved understanding of deep learning; however, building \emph{safe} human-level AI is a harder problem that will likely require a more robust theory.
Thus advancing the science of deep learning seems somewhat more critical for solving problems of safety than of capabilities.
This argument is speculative, and may turn out to be wrong.

\section{Additional Figures}

\begin{figure}[ht]
\centering
\includegraphics[width=\textwidth]{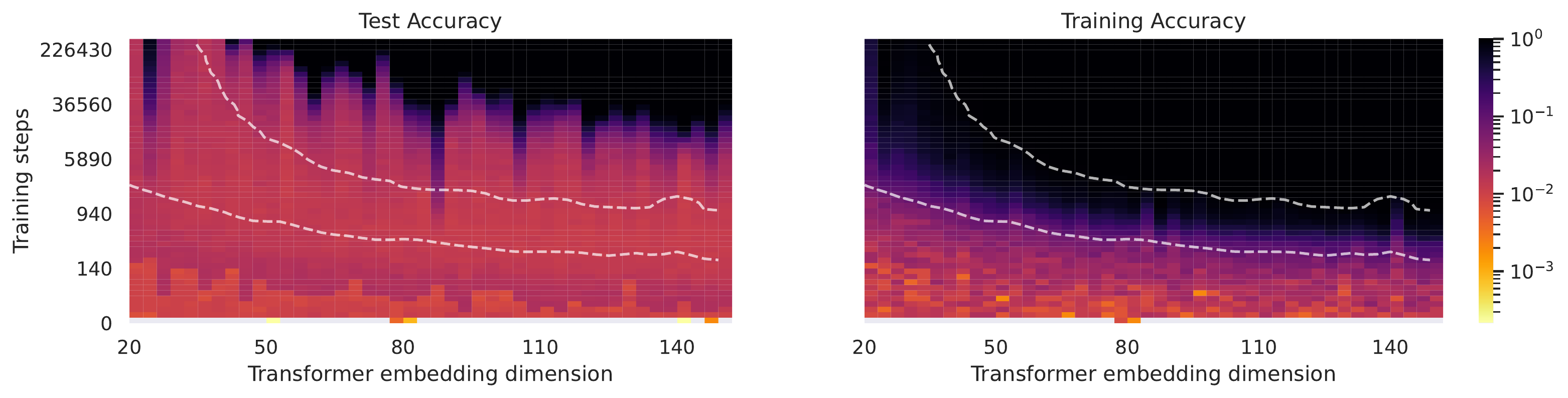}
\caption{
  \textbf{Left:} Test accuracy as function of training steps and model size. A faint double descent in test accuracy is visible, and delayed generalization is clear both epoch-wise (vertically) and model-wise (horizontally).
  \textbf{Right:} Training accuracy as a function of training steps and model size.
}
\label{fig:heatmap-acc}
\end{figure}

\begin{figure}[ht]
\centering
\includegraphics[width=\textwidth]{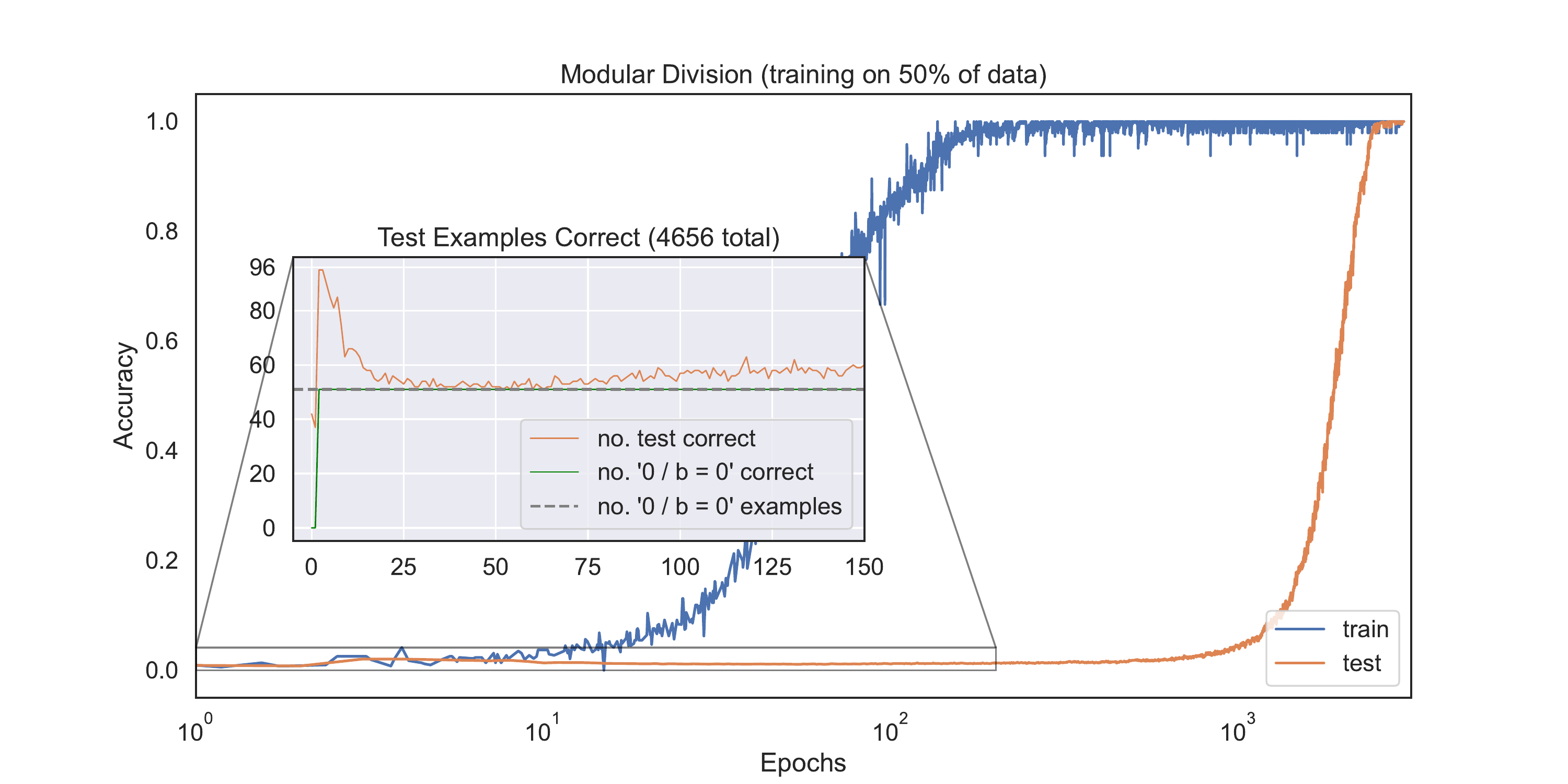}
\caption{
    Early in training, the model learns $0 / b = 0 \mod n$, as the no. $0 / b = 0 \mod n$ correct (green) matches all such examples (dashed gray). This leads to a brief peak of 96 correct examples, corresponding to perfect performance on the $0 / b = 0$ and chance performance (1/97) on other data. Test accuracy on all other examples then falls below chance due to memorization.
}
\label{fig:initial-ascent}
\end{figure}

\end{document}